\documentclass[letterpaper, 10 pt, conference]{ieeeconf}  

\usepackage{graphicx}
\usepackage{amssymb}
\usepackage{amsmath}
\usepackage{commath}
\usepackage{mathtools}

\usepackage[hidelinks]{hyperref}
\usepackage{import}
\usepackage{paralist}
\usepackage{gensymb}
\usepackage{dsfont}
\usepackage{siunitx}
\usepackage[bottom]{footmisc}
\usepackage{framed}
\usepackage{balance}
\usepackage{color, colortbl}
\usepackage[dvipsnames]{xcolor}
\usepackage[skins]{tcolorbox}
\usepackage{url}
\usepackage{cleveref}
\usepackage{cite}
\usepackage{microtype}
\usepackage{tikzscale} 
\usepackage{pgfplots}
\usepackage{eso-pic}
\usepackage{graphicx}

\usepackage{todonotes}

\usepackage{textcomp}
\usepackage{gensymb}
\usepackage{subfig}
\usepackage{booktabs}
\usepackage{multirow}

\usepackage{pifont}
\newcommand{\cmark}{\ding{51}}%
\newcommand{\xmark}{\ding{55}}%

\newsavebox{\measurebox}

\usepackage{amsmath,mathtools}

\DeclareMathOperator*{\argmin}{arg\,min}

\newtcolorbox{myframe_bak}[2][]{%
  enhanced,colback=white,colframe=black,coltitle=black,
  sharp corners,boxrule=0.4pt,left=0pt,right=0pt,top=0pt,bottom=0pt,
  attach boxed title to top left={yshift=-0.3\baselineskip-0.4pt,xshift=2mm},
  boxed title style={tile,size=minimal,left=1mm,right=1mm,
  colback=white,before upper=\strut},
  title=#2,#1
}

\newtcolorbox{myframe}[2][]{%
  enhanced,colback=white,colframe=black,coltitle=black!75!white,
  left=0pt,right=0pt,top=0pt,bottom=0pt,
  attach boxed title to top left={yshift=-0.4\baselineskip-0.4pt,xshift=1.5mm},
  boxed title style={tile,size=minimal,left=1mm,right=1mm,
  colback=white,before upper=\strut},
  title=#2,#1
}

\IEEEoverridecommandlockouts                              

\title{\LARGE \bf
Learning to Walk and Fly with Adversarial Motion Priors
}

\AddToShipoutPictureBG*{%
  \AtPageUpperLeft{%
    \setlength{\unitlength}{1mm}%
    \put(0,-12){\makebox(\paperwidth,0)[c]{\parbox{0.8\textwidth}{\centering\textcolor{gray}{\large This paper has been accepted for publication at the IEEE/RSJ International Conference on Intelligent Robots and Systems (IROS), Abu Dhabi, 2024.  ©IEEE}}}}
  }
}

\author{Giuseppe L'Erario$^{1,3}$, Drew Hanover$^{2}$, Ángel Romero$^{2}$, Yunlong Song$^{2}$, \\Gabriele Nava$^{1}$, Paolo Maria Viceconte$^{1}$, Daniele Pucci$^{1,3}$, Davide Scaramuzza$^{2}$
\thanks{$^{1}$Artificial and Mechanical Intelligence, Istituto Italiano di Tecnologia, Genoa, Italy
$^{2}$Robotics and Perception Group, University of Zurich, Switzerland.
$^{3}$University of Manchester, Manchester, UK. 
D. Hanover, A. Romero, Y. Song, and D. Scaramuzza were supported by the European Research Council (ERC) under grant agreement No. 864042 (AGILEFLIGHT).
}
}

\begin{document}

\maketitle
\thispagestyle{empty}
\pagestyle{empty}
\everypar{\looseness=-1}

\begin{abstract}


  Robot multimodal locomotion encompasses the ability to transition between walking and flying, representing a significant challenge in robotics. This work presents an approach that enables automatic smooth transitions between legged and aerial locomotion. Leveraging the concept of Adversarial Motion Priors, our method allows the robot to imitate motion datasets and accomplish the desired task without the need for complex reward functions. The robot learns walking patterns from human-like gaits and aerial locomotion patterns from motions obtained using trajectory optimization.
  Through this process, the robot adapts the locomotion scheme based on environmental feedback using reinforcement learning, with the spontaneous emergence of mode-switching behavior.
  The results highlight the potential for achieving multimodal locomotion in aerial humanoid robotics through automatic control of walking and flying modes, paving the way for applications in diverse domains such as search and rescue, surveillance, and exploration missions.
  This research contributes to advancing the capabilities of aerial humanoid robots in terms of versatile locomotion in various environments.
  \newline
  \textbf{Video}: \url{https://youtu.be/mi6Do-x67CM}
\end{abstract}

\newcommand{\asFrame}[1]{\mathcal{#1}}

\newcommand{\R}{\mathbb{R}}

\newcommand{\com}{x_\text{CoM}}
\section{Introduction}

The transition between locomotion styles is a phenomenon that many species across the animal kingdom exhibit.
Not many robots, however, can combine aerial and terrestrial locomotion, thus leaving the problem of traversing
different environments still partially open. The broader research community has begun to study what it takes to enable multimodal locomotion for various bio-inspired robotic systems \cite{lock2013multi}.
Countless examples have demonstrated systems with bespoke forms of locomotion: bat-inspired robots that can fly and walk~\cite{daler2015bioinspired},
propeller-driven bipedal robots~\cite{kim2021bipedal}, morphing rovers~\cite{sihite2023multi}, rolling quadrupeds~\cite{bjelonic2019keep}, underwater robots~\cite{yu2023multimodal, yu2011design} and even flying humanoid robots~\cite{Pucci2017MomentumRobot, Nava2018PositionRobot}.

However, little work exists on understanding how and when to transition between the locomotion forms.
For example, when dealing with an aerial-humanoid robot, what is the proper time to transition between modes such as walking or flying when considering a high-level locomotion task?
This goal can be accomplished by breaking out the controller into submodules, where each module handles an individual locomotion task, as proposed by~\cite{kim2021bipedal}.
Nevertheless, the transition between locomotion forms is a significant challenge and remains an open question.
This paper moves forward with a learning-based method that enables
aerial humanoid robots to exhibit multimodal locomotion capabilities. \looseness=-1

A standard approach to address the locomotion problem is to decompose the system into two layers~\cite{romualdi2020benchmarking}:
1) A trajectory optimization (TO) layer, whose role is to provide a set of feasible trajectories~\cite{Kelly2017AnCollocation, dafarra2022dynamic};
2) Whole-body control, which stabilizes the trajectories produced by the TO layer~\cite{kuindersma2016optimization}.

\begin{figure}
    \centering
    \includegraphics[width=1.0\linewidth]{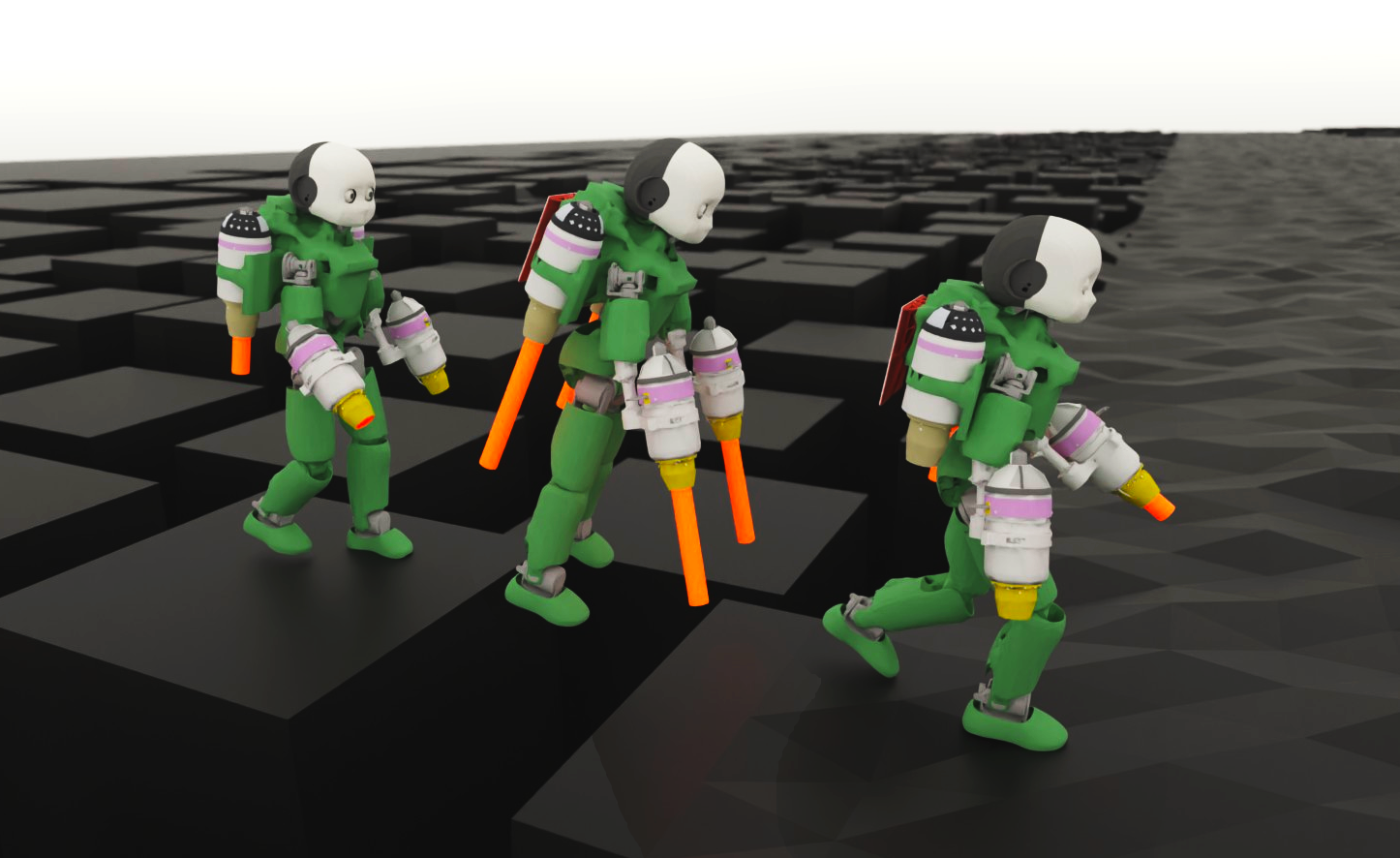}
    \caption{iRonCub, the aerial humanoid robot, on a complex terrain.} 
    \label{fig:ironcub}
    \vspace{-0.5cm}
\end{figure}

The TO layer lies at the core of the locomotion architecture, and it is responsible for generating a set of feasible trajectories that the robot can track. It is formulated as an optimization problem that minimizes a cost function -- encoding the desired robot behavior -- subject to constraints -- characterizing the system dynamics, also called \textit{model}.
Typical approaches for legged systems as humanoid robots rely on \textit{simplified} models such as the linear inverted pendulum (LIP)~\cite{kajita2003biped}, or the divergent component of motion (DCM)~\cite{englsberger2013three}.
While optimization methods using these simplified models are fast, they cannot exploit the robot's complete structure. In contrast, approaches leveraging the robot's \textit{full} dynamics can utilize its inherent system structure, albeit they are computationally expensive.
\textit{Reduced} models as the \textit{centroidal momentum dynamics}
are a compromise between the two extremes and reduce the problem's complexity~\cite{orin2013centroidal,Dai2014, Ponton2018OnDynamics,romualdiCentroidalMPC}.
Nonetheless, TO methods remain computationally intensive and demand prior knowledge of the environment. On the other side of the spectrum, model-free techniques such as Reinforcement Learning (RL) can learn a policy that maps the state of the system to action without the need for a model~\cite{SuttonRichardSBartoReinforcementLearning}. These tools enable a new set of control strategies for the robotics realm~\cite{tsounis2020deepgait, yunlong_science, rodriguez2021deepwalk, li2021reinforcement, kaufmann2023champion}. The main challenge in RL for robot control lies in the limited ability of trained agents to exhibit natural behavior in specific tasks, often resulting in inefficient solutions that exploit simulator inaccuracies.
To address this limitation, researchers have focused on how nature's ingenuity has shaped the movements of animals, such as motor skills exhibited by humans, which they embed into simulated through data-driven approaches~\cite{peng2018deepmimic, peng2021amp, peng2022ase, starke2020local, starke2022deepphase}.
These techniques recently found their way into robotics, where they are used to train robots to imitate animal-like motions~\cite{escontrela2022adversarial, RoboImitationPeng20, viceconte_adherent}.
One such method, called Adversarial motion priors (AMP)~\cite{peng2021amp}, takes cues from Generative Adversarial Imitation Learning (GAIL) to train an agent to replicate the ``style'' embedded in a reference dataset~\cite{ho2016generative}.

This paper progresses in this direction by proposing a method that utilizes AMP and RL to study the spontaneous emergence of automatic and smooth locomotion transitions in aerial humanoid robots.
AMP allows the robot to learn and mimic human-like motions when walking - enhancing the naturalness of the gait - and imitating flying motion obtained via trajectory optimization.
The output of the proposed approach is a locomotion pattern that switches between human-like and flying locomotion based on the task and the environment.
We train locomotion policies for iRonCub, a jet-powered aerial humanoid robot with terrestrial and aerial locomotion capabilities.
We learn to mimic the motion styles provided within the motion datasets while achieving a high-level waypoint tracking task.
The transition between flying and walking is managed by including an energy proxy term in the reward function, encouraging the robot to walk when the ground is within reach and fly otherwise. The method is tested in both the cases of \textit{ideal} thrust propulsion and \textit{specific} jet-powered actuation modeled from real-world data.

We performed experiments within the Nvidia Isaac Gym environment~\cite{makoviychuk2021isaac,rudin2022learning}, forcing the agent to traverse complex terrains and rough courses.
To our knowledge, this is the first demonstration of an aerial humanoid robot exhibiting smooth transitions between walking and flying without explicitly tracking a trajectory provided by trajectory optimization or using a state machine to switch between locomotion modes.
Instead, our agent learns how to fly, walk, and navigate without explicitly being told how or what locomotion to use.

The paper is organized as follows. Sec~\ref{sec:background} introduces notation and recalls some critical ideas about floating-base systems. Sec~\ref{sec:method} presents the whole approach by recalling the AMP method and introducing the system modeling. Sec~\ref{sec:results} validates the technique on iRonCub, a flying humanoid robot. Sec~\ref{sec:conclusions} concludes the paper with remarks and future work.






\section{Background}
\label{sec:background}


\subsection{Floating-base modeling}
\label{sec:floating-base}

A robot can be modeled as a multi-body system composed of $n+1$ rigid bodies -- called links -- connected by $n$ joints with one degree of freedom. Using of the \textit{floating base} formalism~\cite{traversaro2017modelling}, the robot's configuration is defined by the tuple $q = ({}^\asFrame{I}p_\asFrame{B}, {}^\asFrame{I}R_\asFrame{B}, \theta) \in \mathbb{Q}$, where ${}^\asFrame{I}p_\asFrame{B} \in \R^3$ and ${}^\asFrame{I}R_\asFrame{B} \in SO(3)$ are the position and orientation of the robot base frame $\asFrame{B}$ w.r.t. the inertial frame $\asFrame{I}$, and $\theta \in \R^n$ are the joint positions.
The element $\nu = ({}^\asFrame{I}\dot{p}_\asFrame{B}, {}^\asFrame{I}\omega_\asFrame{B}, \dot{\theta})$
is the robot's velocity, where $\dot{\theta}$ are the joint velocities and $({}^\asFrame{I}\dot{p}_\asFrame{B}, {}^\asFrame{I}\omega_\asFrame{B})$ are the linear and angular velocity of the base frame $\asFrame{B}$ w.r.t. $\asFrame{I}$ such that $\dot{{}^\asFrame{I}R_\asFrame{B}} = S({}^\asFrame{I}\omega_\asFrame{B}) {}^\asFrame{I}R_\asFrame{B} $ is satisfied.
By applying the Euler-Poincaré formalism, the equation of motion of a system exchanging $n_b$ wrenches with the environment results in
\begin{equation}
    M(q) \dot{\nu} + C(q, \nu) \nu + G(q) = \begin{bmatrix} 0_{6 \times 1} \\ \tau \end{bmatrix} + \sum _{i=1} ^{n_b} J^\top _i \mathrm{f}_i,
    \label{system_dyn}
\end{equation}
where $M, C \in \R ^{(n+6) \times (n+6)}$ are the mass and Coriolis matrix, $G \in \R ^{n+6}$ is the gravity vector, $\tau \in \R ^{n}$ are the internal actuation torques, $\mathrm{f}_i$ is the $i$-th of the $n_b$ external wrench applied on the origin of the frame $\asFrame{C}_i$ and $J_i \in \R ^{6 \times (n+6)}$ is the jacobian mapping the system velocity $\nu$ to the velocity $({}^\asFrame{I}\dot{p}_{\asFrame{C}_i}, {}^\asFrame{I}\omega_{\asFrame{C}_i})$ of the frame $\asFrame{C}_i$.
These quantities are useful to understand the choice of the state and action space of the robot. \looseness=-1

\subsection{Reinforcement Learning}
\label{sec:rl}

We model the problem as a Markov Decision Process (MDP) defined by a tuple $(\mathcal{S}, \mathcal{A}, \mathcal{P}, \mathcal{R}, \gamma)$~\cite{SuttonRichardSBartoReinforcementLearning}. The state space $\mathcal{S}$ is the set of all possible states of the system and the environment, while the action space $\mathcal{A}$ is the set of all possible actions that the agent can execute. The transition probability $\mathcal{P}: \mathcal{S} \times \mathcal{A} \times \mathcal{S} \rightarrow \mathsf{Pr}[\mathcal{S}]$ defines the probability of transitioning from a state $s_t$ to a state $s_{t+1}$ given an action $a_t$.
The discount factor $\gamma \in [0, 1)$ trades off
long-term against short-term rewards. In the context of Deep RL, the agent is modeled as a neural network policy $\pi_\sigma(a_t|s_t)$ that maps a state $s_t$ to an action $a_t$, with $\sigma$ the parameters of the neural network. The goal of DRL is to find the optimal set of parameters $\sigma$ that maximize the expected discounted return
\begin{equation}
    J(\sigma)=\mathbb{E}_{\pi_\sigma} \left[\sum_{t=0}^{T-1} \gamma^t r_t\right],
\end{equation}
where $\mathcal{R}(s_t, a_t, s_{t+1})$ represents the reward $r_t$ received when the agent executes an action $a_t$ in $s_t$ and transitions to $s_{t+1}$.


\section{Method}
\label{sec:method}

Our work aims to train a policy $\pi$ capable of walking and flying, automatically selecting the best locomotion pattern to accomplish a task without needing a high-level planner. The agent should be able to imitate the locomotion patterns extracted from a set of motion datasets $\mathcal{D}$, in which each sample represents a snapshot capturing the system's kinematics at a specific instant.


\subsection{Adversarial Motion Priors}
\label{sec:amp}

\begin{figure}
    \centering
    \includegraphics[width=\columnwidth]{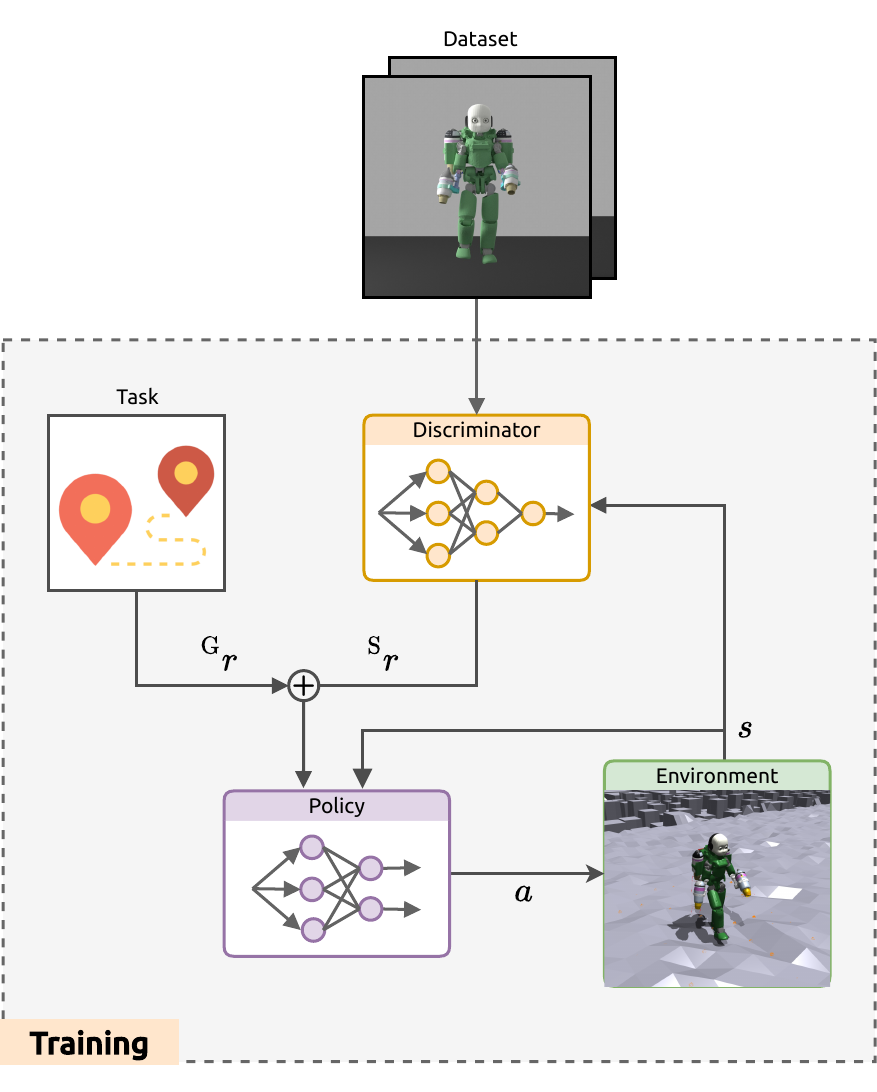}
    \caption{The discriminator learns to distinguish between samples from the dataset and samples produced by the agent. The policy $\pi_\sigma$ is trained to imitate the dataset's motion and accomplish a task simultaneously by maximizing the total reward $r(t)$ that expresses the quality of the motion and the task accomplishment.}
    \label{fig:amp}
\end{figure}

The agent is trained to imitate the motion of the dataset and to accomplish a task at the same time. This requirement translates into a reward function composed of two terms, weighted by scalar values $w_{\text{G}}, w_{\text{S}}$, as proposed in AMP~\cite{peng2021amp}:
\begin{equation}
    r_t = w_{\text{G}} {}^{\text{G}} r_t + w_{\text{S}} {}^{\text{S}} r_t.
\end{equation}

The \textit{task reward} ${}^{\text{G}} r$ specifies the goal the agent should accomplish, e.g., reach a target or have a desired base velocity. The \textit{style reward} ${}^{\text{S}} r$ enforces the policy to produce locomotion patterns resembling the motion of the dataset. The total reward encourages the agent to imitate a desired set of motion patterns while choosing the one that best accomplishes the desired goal. Given the environment, no high-level planner is required to select a specific motion, and the switching behavior emerges naturally. An overview of the system is given in Fig.~\ref{fig:amp}.

\subsubsection{Style-reward}
\label{sec:style_reward}

Generating data that resemble existing information is a characteristic of Generative Adversarial Networks (GANs)~\cite{goodfellow2014generative}. GANs consist of two competing neural networks: a generator that creates new data that resembles the training set and a discriminator that tries to distinguish between real and generated data, trained in a game-like fashion.
Similarly, we can use a discriminator $D$ to distinguish between the samples of the dataset and those produced by the agent $\pi_\sigma$ that, in turn, acts as a generator aiming at creating realistic movements.

In practice, the discriminator $D_\Phi: \mathbb{R}^k \times \mathbb{R}^k \rightarrow \mathbb{R}$ is a neural network with parameters $\Phi$ that maps a couple of consecutive samples $\chi_t, \chi_{t+1}$ of dimension $k$ to a scalar value and aims at differentiating motions from the dataset $\mathcal{D}$ from movements produced by the RL agent following the policy $\pi_\sigma$.
The objective proposed in~\cite{peng2021amp, peng2022ase} to train the discriminator is \looseness=-1
\begin{equation}
    \begin{split}
        \argmin_\Phi \ & -\mathbb{E}_{(\chi_t, \chi_{t+1})\sim\mathcal{D}}
        [\log(D_\Phi(\chi_t, \chi_{t+1}))] \\
        & - \mathbb{E}_{(\chi_t, \chi_{t+1})\sim\mathcal{\pi_\sigma}}[\log (1 - D_\Phi(\chi_t, \chi_{t+1}))] \\
        & + w_{\text{gp}} \mathbb{E}_{(\chi_t, \chi_{t+1})\sim\mathcal{D}} \big [ || \nabla_\Phi D_\Phi(\chi_t, \chi_{t+1})||^2],
    \end{split}
\end{equation}
where the first two terms
force the discriminator to output $D_\Phi(\chi_t, \chi_{t+1})=1$ when the discriminator is fed with dataset transitions and $D_\Phi(\chi_t, \chi_{t+1})=0$ when it is fed with samples produced by the agent. The final term is a gradient penalty that penalizes nonzero gradients on dataset transitions, resulting in improved stability and quality of the GAN training~\cite{mescheder2018training}.

When fed with samples produced by the agent, the discriminator serves as a critic that evaluates the quality of the motion produced by the policy. The output of $D(\chi_t, \chi_{t+1})$ is hence used to compute the style reward~\cite{peng2021amp, peng2022ase}
\begin{equation}
    {}^\text{S} r_t = -\log{\Big(1 - \dfrac{1}{1 + e^{-D_\Phi(\chi_t, \chi_{t+1})}}\Big)}, \label{eq:style_reward}
\end{equation}
in which the negative logarithm is used to map the output of the discriminator to a reward that is high when the agent produces a transition similar to the dataset.



\subsubsection{Task-reward}
\label{sec:task_reward}

We want the robot to reach a target and follow a route made of checkpoints. Ideally, we could reward the agent when it hits the target. This approach makes associating a reward with an individual action complex since it introduces reward sparsity. A common practice is to use proxy rewards that guide the agent to the true objective and provide continuous feedback to the agent. We use three rewards to accomplish the task.

At each time instant, the agent can be associated with a checkpoint. First, we want to minimize the distance of the robot base to the target at each time instant $t$. This requirement translates into the reward:
\begin{equation}
    {}^c r_t = \exp(-c_1 \| x_d - p_t\|^2),
    \label{eq:reward_checkpoint}
\end{equation}
where $x_d \in \mathbb{R}^3$ is the target position, $p_t \in \mathbb{R}^3$ is the position of the robot base, and $c_1$ a hyperparameter.

Second, we want the robot to approach the checkpoint with a desired velocity $v_d \in \mathbb{R}$. This reward is written as:
\begin{equation}
    {}^v r_t = \exp\left(-c_2 \left\lVert v_d - \frac{g(p_t) - g(p_{t-1})}{\Delta t}\right\lVert^2\right),
    \label{eq:reward_velocity}
\end{equation}
where $g(p) = \| p - x_d \|$ is a function that returns the scalar distance between the robot base and the target, $\Delta t$ is the time step, and $c_2$ is a hyperparameter. This term encourages the agent to travel at a desired velocity $v_d$ on a line connecting the robot base to the target.

The last reward demands the agent to face the checkpoint, i.e., the robot base $x$-axis should point to the projection of the target in the horizontal plane:
\begin{equation}
    {}^f r_t = \min(0, f(p_t)_{xy} \cdot {}^\asFrame{I}i_{xy}),
    \label{eq:reward_facing}
\end{equation}
where $f(p) = \frac{x_d - p}{\| x_d - p \|}$ is a function that returns the unit vector pointing from the robot base to the target, the subscript $xy$ extracts the horizontal components, and the unit vector ${}^\asFrame{I}i$ is the robot $x$-axis in the inertial frame.

The reward components~\eqref{eq:reward_checkpoint},~\eqref{eq:reward_velocity}, and~\eqref{eq:reward_facing} fully describe the task and become equal to $1$ when the robot reaches the target.

Additionally, we want to minimize thrust usage. The thrust penalty can be considered a proxy term that minimizes propulsion expenditure and encourages the agent to use legged locomotion when possible. The thrust penalty is written as
\begin{equation}
    {}^{\text{T}} r_t = -\left\lVert\frac{T_t}{T_\text{max}} \right\lVert^2 ,
    \label{eq:reward_thrust}
\end{equation}
where $T_\text{max}$ is the maximum thrust each jet can exert while $T_t \in \mathbb{R}^m$ is the thrust vector acting on the system.

Finally, the total task reward is written as
\begin{equation}
    {}^{\text{G}}r_t = w_c {}^c r_t + w_v {}^v r_t + w_f {}^f rt + w_{\text{T}} {}^{\text{T}}r_t,
    \label{eq:reward_task}
\end{equation}
where $w_c$, $w_v$, $w_f$, and $w_{\text{T}}$ are hyperparameters that weight the contribution of each reward.

\subsection{Model representation}
\label{sec:action_observation_space}

Sec~\ref{sec:floating-base} recalls how the quantities describing a multi-body system affect the robot's evolution. We build the action and the observation spaces accordingly, considering also the environment and task information.

\subsubsection{Observation Space}

The observation space consists of two components: one relates to the robot, and one distillates information about the task and the environment. We define the robot observation vector as
\begin{equation}
    o_{\text{robot}} = \begin{bmatrix}
        \rho^\top                             &
        {}^\asFrame{B}v_\asFrame{B}^\top      &
        {}^\asFrame{B}\omega_\asFrame{B}^\top &
        \theta^\top                           &
        \dot{\theta}^\top                     &
        p_\text{EE}^\top                      &
        T^\top
    \end{bmatrix}^\top,
    \label{eq:robot_observation}
\end{equation}
where $\rho$ is the quaternion representation of the rotation of the base ${}^\asFrame{I}R_B$, ${}^\asFrame{B}v_\asFrame{B}$ and ${}^\asFrame{B}\omega_\asFrame{B}$ are the base velocity in the base frame, $\theta$ and $\dot{\theta}$ are the joints position and velocity, $p_\text{EE}$ is the vector containing the relative position of the end effectors, namely hands and feet, and $T = [T_1, \dots, T_m]$ is the thrust acting on the system as specified in Eq.~\eqref{eq:thrust-force}.

The environment observation gives information about the robot's interaction with the environment. We use an elevation map simulating a scan of the terrain around the robot
\begin{equation}
    {o}_\text{env} = \begin{bmatrix}
        h_{11}        & h_{12}        & \dots  & h_{0\text{L}} \\
        h_{21}        & h_{22}        & \dots  & h_{2\text{L}} \\
        \vdots        & \vdots        & \ddots & \vdots        \\
        h_{\text{V}1} & h_{\text{V}2} & \dots  & h_{\text{VL}}
    \end{bmatrix}.
    \label{eq:env_observation}
\end{equation}

The elevation map is centered at the robot base and moves with the robot. The grid is discretized into $\text{V} \times \text{L}$ cells. $h_{ij} \in \mathbb{R}$ is the height of the $i$-th row and $j$-th column of the grid map with respect to the robot base.

The task observation vector gives the agent information about the goal. We use the position of the target $x_d$ with respect to the robot base $p$
\begin{equation}
    o_{\text{task}} = {}^\asFrame{B}p_d = {}^\asFrame{I}R_\asFrame{B}^\top (x_d - p).
    \label{eq:task_observation}
\end{equation}

The observation space is the concatenation of the robot, environment, and task observations
\begin{equation}
    o = \begin{bmatrix}
        o_{\text{robot}}^\top     &
        \bar{o}_{\text{env}}^\top &
        o_{\text{task}}^\top
    \end{bmatrix}^\top,
    \label{eq:observation_space}
\end{equation}
where $\bar{o}_{\text{env}}$ is the flattened environment observation matrix.



\subsubsection{Action Space}

The action space $\mathcal{A}$ is composed of the desired joint positions of the robot $\theta_d \in \mathbb{R}^{n}$ and the desired thrust dynamics input $u = [u_1, \dots, u_m]^\top \in \mathbb{R}^{m}$
\begin{equation}
    a = \begin{bmatrix}
        \theta_d^\top &
        u^\top
    \end{bmatrix}^\top,
    \label{eq:action_space}
\end{equation}
where $n$ is the number of joints of the robot and $m$ is the number of links on which the thrust acts. The desired joint positions $\theta_d$ are fed to a PD controller. The resulting thrust intensities $T$ are computed as the increment of the thrust intensity with respect to the previous time step $k-1$ as ${T = T[k-1] + g(T[k-1], u) \Delta t}$ and act on the specified links, generating a pure force
\begin{equation}
    f_i = {}^\asFrame{I}R_i(q) \begin{bmatrix}
        0 \\
        0 \\
        T_i
    \end{bmatrix}, \forall i \in [1, m],
    \label{eq:thrust-force}
\end{equation}
where ${}^\asFrame{I}R_i(q)$ represents the orientation of the frame $i$ on which the force $f_i$ is applied and $\Delta t$ is the time step between two consecutive actions. $g(T[k-1], u)$ is a function that computes the thrust dynamics given the previous thrust intensities $T[k-1]$ and the input $u$, which in the \textit{ideal} case is considered to be $g(T[k-1], u) = \dot{T}$, i.e., thrust intensity rate-of-change, while in the \textit{specific} use case of the jet-powered actuation, it is a function identified from real-world data, see Sec.~\ref{sec:jet_propulsion} for more details.

\subsection{Discriminator Observation Space}
\label{sec:discriminator_observations}

The discriminator $D$ gives feedback about the quality of the motion produced by the agent. Selecting a meaningful set of features to feed into the discriminator is crucial since it should be able to capture the robot's motion. We choose the discriminator observation space to be the same as the robot observation vector $\chi = o_{\text{robot}}$. The discriminator is fed with couples of consecutive observations $\chi_t, \chi_{t+1}$ and trained with batches of samples from the motion dataset $\mathcal{D}$ and samples produced by the policy $\pi_\sigma$.

\section{Results}
\label{sec:results}

In this section, we present the simulated results of the proposed approach.
We designed our experiments to answer the following questions: (i) Can smooth multimodal locomotion be achieved with an AMP-based method? (ii) What type of information does the policy need to guide the emergence of the desired locomotion pattern? (iii) Can the same result be obtained using a classic RL approach?

Our testbench is iRonCub, a flying humanoid robot that expresses a degree of terrestrial and aerial locomotion. iRonCub has 23 joints, weights $44~\si{kg}$, and is equipped with four jet engines,
two fixed on the chest and two moved by the arms, which in the specific use case are the commercial JetCat P250 engines~\cite{JetCatP250}, capable of exerting $T_{max}=250~\si{N}$ of thrust each.

The training consists of two main tasks. The first task is a multimodal locomotion scenario on flat ground, in which the robot has to catch several waypoints located on the ground and in the air. The second scenario consists of a world containing different terrains, in which the robot has to understand which locomotion modality is more suited. While the scenarios above are tested in the \textit{ideal} thrust case, we also validate the approach in the \textit{specific} use case of jet-powered actuation, in which the model of the jet engines is identified from real-world data with a neural network and embedded in the simulation environment, see Sec.~\ref{sec:jet_propulsion}.

\subsection{Motion Priors}
\label{sec:motion_priors}

\begin{figure}
    \centering
    \subfloat{
        \includegraphics[width=0.3\linewidth, trim={0 0cm 0 0cm},clip]{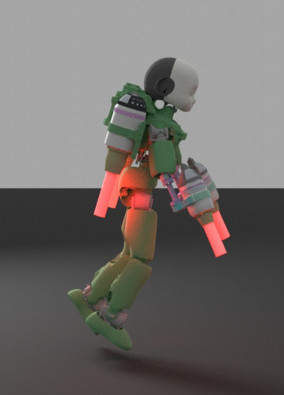}} \hfill
    \subfloat{
        \includegraphics[width=0.3\linewidth, trim={0 0cm 0 0cm},clip]{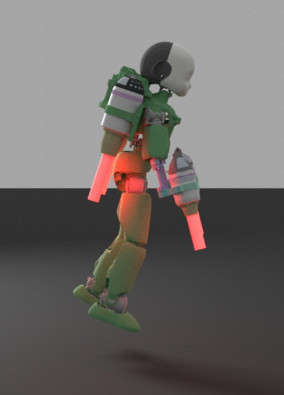}} \hfill
    \subfloat{
        \includegraphics[width=0.3\linewidth, trim={0 0cm 0 0cm},clip]{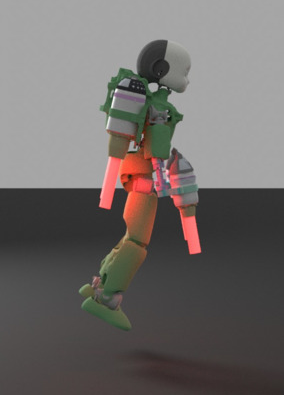}}
    \caption{Snapshots of the flying motion obtained using TO.}
    \label{fig:fly-motion}
    \centering
    \subfloat{
        \includegraphics[width=0.3\linewidth, trim={0 0cm 0 0cm},clip]{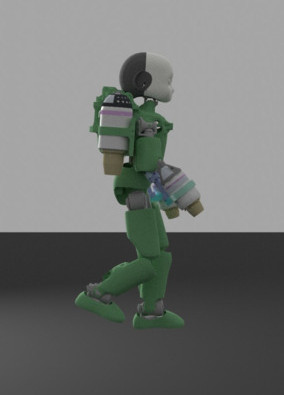}} \hfill
    \subfloat{
        \includegraphics[width=0.3\linewidth, trim={0 0cm 0 0cm},clip]{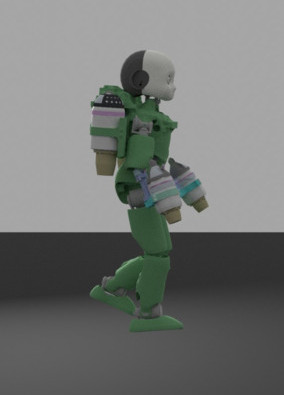}} \hfill
    \subfloat{
        \includegraphics[width=0.3\linewidth, trim={0 0cm 0 0cm},clip]{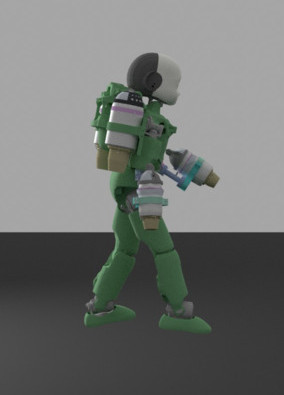}}
    \caption{Snapshots of the walking motion obtained using inverse kinematics from the CMU human walking dataset.}
    \label{fig:walk-motion}
\end{figure}
The used motion data comes from two sources. (i) Walking motions are produced from recorded mocap clip~\cite{CMU-dataset, peng2018deepmimic} and retargeted to the iRonCub model using inverse kinematics implemented using \textit{iDynTree}~\cite{10.3389/frobt.2015.00006} (Fig.~\ref{fig:walk-motion}). (ii) Trajectory optimization for aerial humanoid robots produces flying motions containing the full state~\cite{lerario2022multimodal}. These datasets consist of motions described in terms of the system's kinematics, i.e., joint positions, base poses, and thrust intensities (Fig.~\ref{fig:fly-motion})
In the case of walking motions, the thrust is set to zero.
If the original demonstrator is different from the agent, the motion might not be feasible. In this case, the agent will produce a motion similar to the original one but feasible for the agent.

\begin{table}[t]
    \caption{PPO parameters.}
    \centering
    \begin{tabular}{cc}
        \midrule
        Parameter               & Value \\
        \midrule
        \rowcolor{gray!15}
        Discount rate $\gamma$  & 0.99  \\
        Learning rate           & 5e-5  \\
        \rowcolor{gray!15}
        GAE parameter $\lambda$ & 0.95  \\
        Entropy coefficient     & 0.0   \\
        \rowcolor{gray!15}
        Clip parameter          & 0.2   \\
        Mini-batch size         & 32768 \\
        \rowcolor{gray!15}
        Critic loss coefficient & 5     \\
        KL threshold            & 0.008 \\
        \rowcolor{gray!15}
        Number of actors        & 4096  \\
        \bottomrule
    \end{tabular}
    \label{tab:PPO_parameters}
\end{table}


\subsection{Experimental Setup}

\begin{figure}
    \centering
    \includegraphics[width=0.49\linewidth]{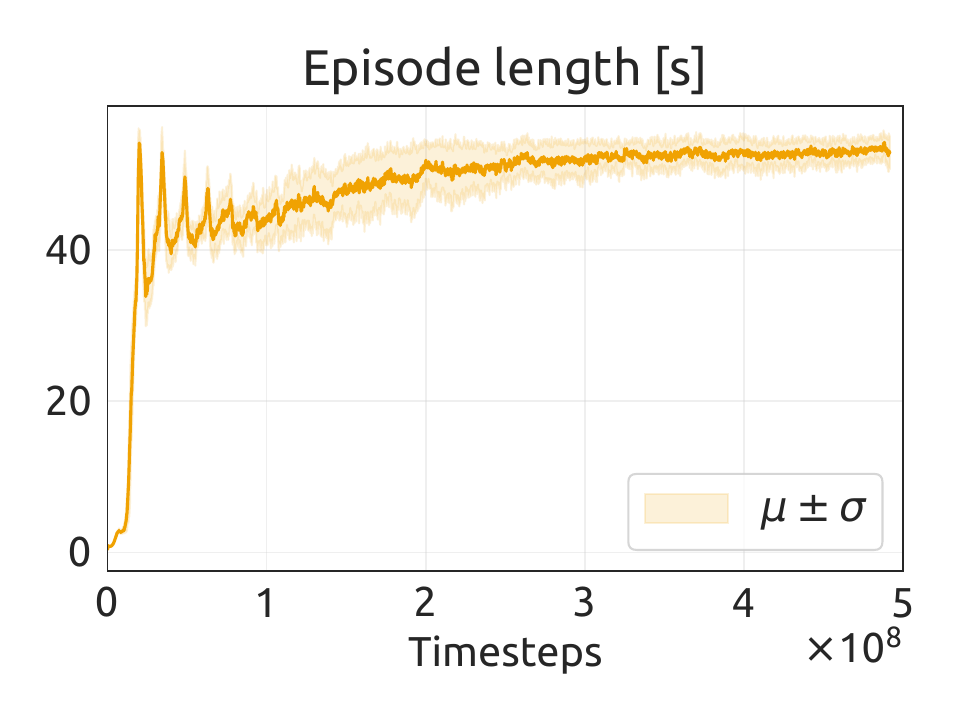}
    \includegraphics[width=0.49\linewidth]{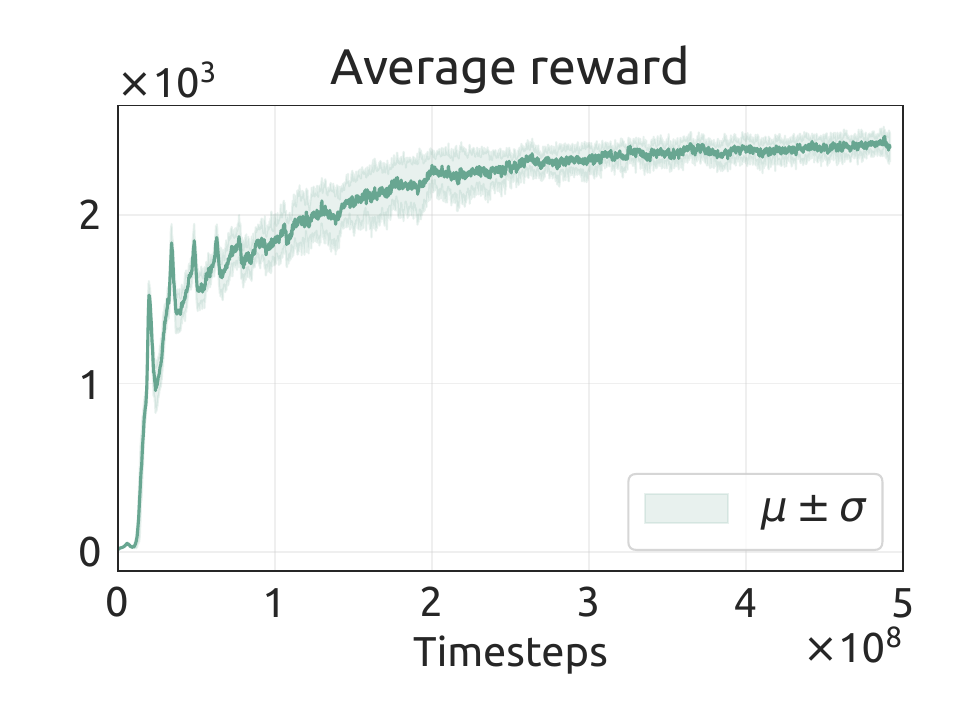}
    \caption{Training curves over 10 runs on the scenario from Sec.~\ref{sec:terrain-aware-locomotion}. }
    \label{fig:learning_curves}
\end{figure}

The training environment is developed using the IsaacGym simulator, which allows massive parallel training. We trained 4096 agents using PPO, controlled at $60~\si{hz}$. The training requires $\sim 2$ hours on an NVIDIA Quadro RTX 6000. The training parameters are shown in Tab.~\ref{tab:PPO_parameters}. Fig.~\ref{fig:learning_curves} shows the learning curves of average reward and episode duration over ten training runs for the test presented in Sec.~\ref{sec:terrain-aware-locomotion}. The curves show a small variance, demonstrating the stability of the training. The reward coefficients are shown in Tab.~\ref{tab:reward_coefficients}.

\subsection{Traversing flat ground}
\label{sec:walking_of_flat_ground}

\begin{figure*}
    \centering
    \includegraphics[width=1.0\linewidth, trim={0 1.5cm 0 3cm},clip]{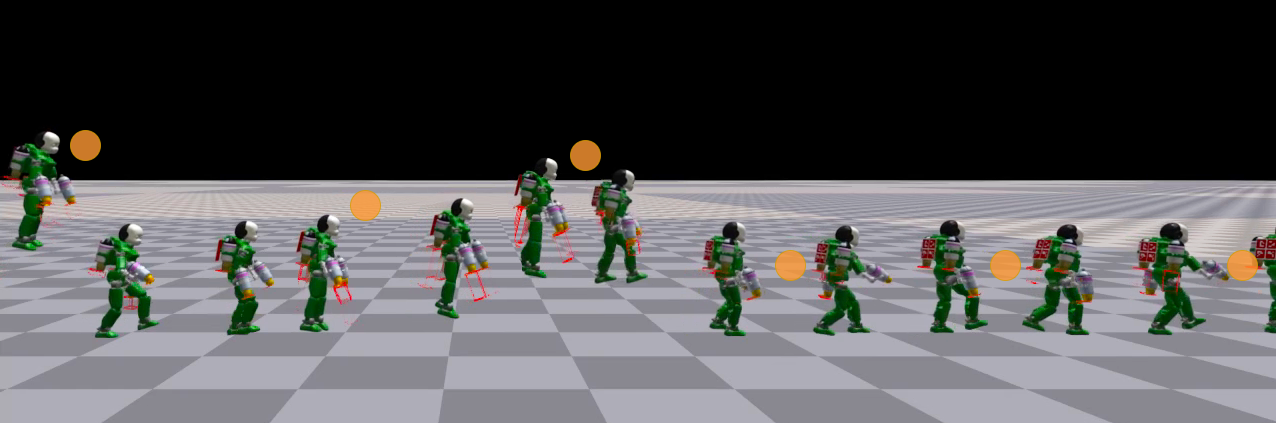}
    \caption{Walk-to-fly maneuver. The robot lands and walks to catch a waypoint on the ground level, takes off, and flies to hit the aerial target.}
    \label{fig:walk-to-fly-transition}
\end{figure*}

The first scenario involves the robot walking, take-off, flying, and landing on flat ground at a desired velocity $v_d$ of $\SI{0.8}{\meter/\second}$. The robot must reach several waypoints set in an interval between $\SI{0.7}{m}$ and $\SI{2.0}{m}$. Once a waypoint is hit, a new one appears forward, and the robot moves toward it, as shown in Fig.~\ref{fig:walk-to-fly-transition}. In this case, the height map
is reduced to a single value, i.e., the height of the robot base w.r.t. the terrain. The test shows that the robot effectively learns terrestrial and aerial locomotion and can choose when to switch between them. Each episode is terminated when the distance of the robot base from the ground is less than $0.4~\si{m}$ or the maximum number of steps is reached. \looseness=-1

\begin{figure*}
    \centering
    \includegraphics[width=1.0\linewidth, trim={0 1.5cm 0 3cm},clip]{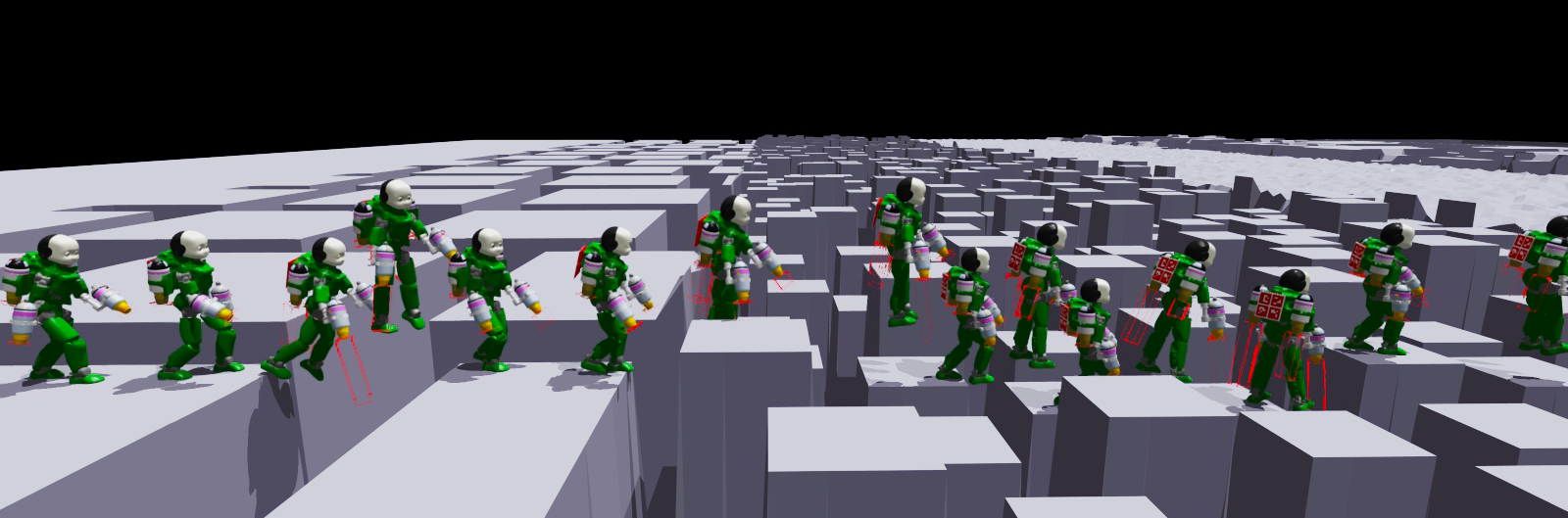}
    \caption{The robot walks over any reachable terrain. The height map enables \textit{terrain-aware} multimodal locomotion.}
    \label{fig:diverse-terrains}
\end{figure*}

\subsection{Terrain-aware locomotion}
\label{sec:terrain-aware-locomotion}

In the second scenario, the robot deals with diverse terrains, i.e., flat and rough terrain, stepping stones, and pits at a desired velocity $v_d$ of $\SI{0.8}{\meter/\second}$. As in the previous scenario, the route comprises waypoints the robot needs to catch. The policy is fed with a height map of $7 \times 9$ cells, where each cell is $\SI{0.3}{m}$ wide.
Using the height map as observation, the robot learns to use small portions of reachable terrain to step on, walk when possible, and fly when the ground is not reachable, as shown in Fig.~\ref{fig:diverse-terrains}.
The termination strategy is the same as the previous case: when the base height is smaller than $0.4~\si{m}$, the episode is over.

Tuning the thrust penalization is not trivial. If the penalty is too high, the robot prefers to stay at the edge of the terrain rather than fly and reach the following reachable terrain since using the thrust leads to a higher penalty than the reward. If the penalty is too low, the robot prefers to fly rather than walk when the terrain is sparser. The penalty should be high enough to make the robot prefer walking over flying but not too high to make the robot choose to walk when flying is needed.  \looseness=-1

\begin{table}[t]
    \caption{Reward coefficients.}
    \centering
    \begin{tabular}{cc}
        \midrule
        Reward term                          & Value \\
        \midrule
        \rowcolor{gray!15}
        Target reward weight $w_c$           & 0.1   \\
        Velocity reward weight $w_v$         & 0.7   \\
        \rowcolor{gray!15}
        Facing reward weight $w_f$           & 0.2   \\
        Thrust penalty weight $w_\text{T}$   & 0.11  \\
        \rowcolor{gray!15}
        Target reward hyperparameter $c_1$   & 0.5   \\
        Velocity reward hyperparameter $c_2$ & 0.5   \\
        \bottomrule
    \end{tabular}
    \label{tab:reward_coefficients}
\end{table}

\subsection{Ablation Studies}
\label{sec:ablation_study}

\begin{figure}
    \centering
    \includegraphics[width=0.7\linewidth]{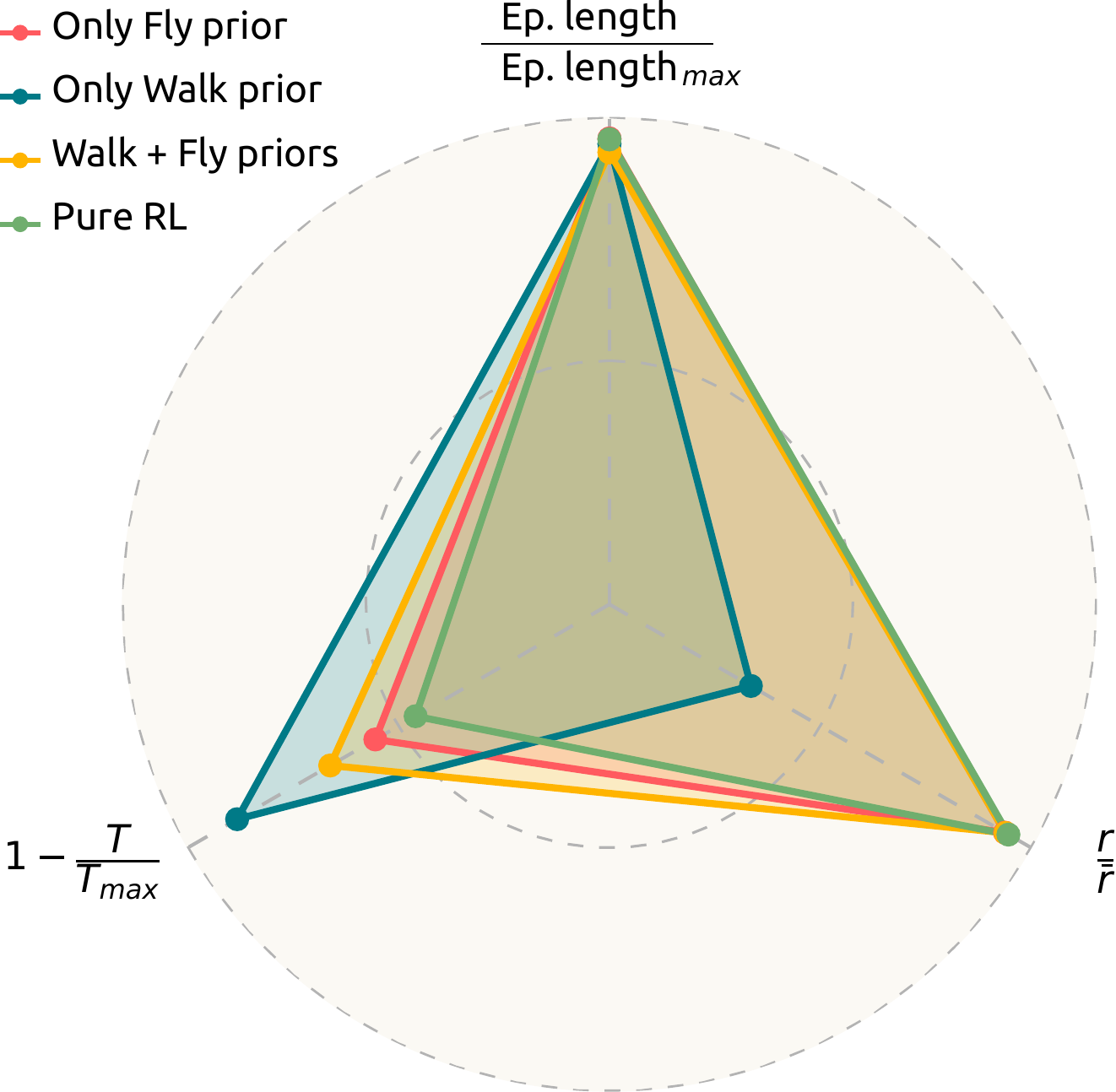}
    \vspace*{0.2cm}
    \caption{The radar chart shows the average quantities in percentage. The policy trained with only walk motion prior shows less usage of the thrust but collects less reward. The other three policies collect higher rewards, with the policy trained with both motion priors -- in yellow -- being efficient in thrust usage and covering a larger area.}
    \label{fig:ablation_study}
\end{figure}

We perform ablation studies to show the influence of the motion priors. We trained the policies using the same parameters but without the motion priors or part of them. We keep the thrust penalization constant and use the scenario described in Sec.~\ref{sec:walking_of_flat_ground} as a testbench. Fig.~\ref{fig:ablation_study} shows episode duration over the maximum duration, reward $r$ collected over a maximum observed reward $\bar{r}$, and thrust usage as $1 - T/T_{max}$. \looseness=-1


\subsubsection{No Motion Prior}
\label{sec:no_motion_prior}

The policy trained without any motion prior does not show any particular behavior. The reward collected is similar to the policy trained with both motion priors, but the thrust usage is higher.

\subsubsection{Only Walking Motion Prior}
\label{sec:no_flying_motion_prior}

The policy trained without aerial motion priors learns how to walk, but it does not show any flying behavior, preventing the robot from reaching the waypoints in the air and collecting rewards. The usage of the thrust is lower than the other policies.

\subsubsection{Only Flying Motion Prior}
\label{sec:no_walking_motion_prior}

The policy trained without terrestrial motion priors does not learn to walk and fly to reach all the waypoints, collecting a reward similar to the other policies. Despite the thrust usage being higher than the policy trained with both motion priors, it is lower than the policy trained without any motion prior~\ref{sec:no_motion_prior}, suggesting that the aerial motion prior embeds an optimized thrust usage coming from the trajectory optimization solution.

\subsubsection{Walking + Flying Motion Prior}

The policy trained with both motion priors learns how to walk and fly. The collected reward is similar to other policies, but the thrust usage is lower, indicating that switching between walking and flying is also more efficient in propulsion expenditure.

\begin{figure*}
    \centering
    \includegraphics[width=1.0\linewidth]{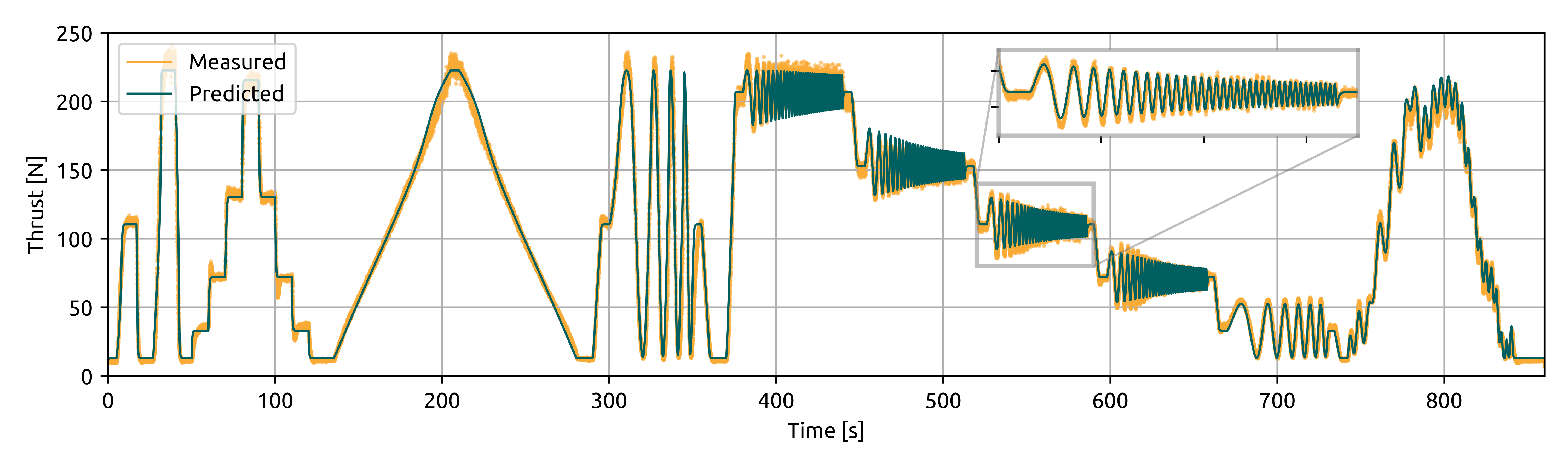}
    \caption{The plot shows the identification results obtained on the JetCat P250. The orange line is the thrust measured from a load cell in a custom test bench for jet data collection. The red line is the thrust obtained by integrating the identified LSTM model.}
    \label{fig:p250-identification-validation}
\end{figure*}

\subsection{Comparison with Trajectory Optimization}
\label{sec:comparison_with_trajectory_optimization}

The trajectory optimization method proposed in~\cite{lerario2022multimodal} is formulated as a multiple-shooting optimal control problem using the centroidal dynamics over an optimized horizon of 100 nodes that requires minimization of thrust usage. The contact sequence is computed using a complementarity condition formulation, and the problem is solved using the IPOPT solver~\cite{wachter2006implementation}.
The TO method is able to produce a transition from legged to aerial locomotion but requires a complex cost function and a long computation time preventing replanning capabilities. A trajectory optimization approach would need the concatenation of several offline trajectories computed for each possible scenario, stabilized afterward by an online controller, while the proposed method can adapt to the environment online. Conversely, the policy can learn how to switch between walking and flying without the need for a complex reward function while inferring the terrain information from the elevation map, which is infeasible with the compared trajectory optimization method or classical control method, needing a separate module that deals with the terrain information. Table~\ref{tab:comparison} compares the two approaches.

\subsection{Use case: Jet-powered actuation}
\label{sec:jet_propulsion}

This section presents the results and considerations in the use case of jet-powered actuation. The jet engines are modeled using a Long Short-Term Memory (LSTM) neural network~\cite{hochreiter1997long}, which is trained to predict the thrust given the input and the state of the jet engine. The model is trained using the data collected from the JetCat P250, a commercial jet engine capable of exerting a thrust of $T_{max} = 250~\si{N}$~\cite{JetCatP250}. The data comprises thrust measured with a load cell at different throttle inputs in an ad-hoc designed test bench~\cite{LErario2020ModelingRobotics}.
The resulting discrete jet dynamics model is
\begin{equation}
    T = T[k-1] + g(T[k-1], u) \Delta t,
    \label{eq:jet-dynamics-lstm}
\end{equation}
where $T$ and $T[k-1]$ are the actual and the previous thrust, $u$ is the throttle input, $g$ is the LSTM model and $\Delta t$ is the time step.
The model is validated by comparing the thrust obtained by integrating the model against the thrust measured from the load cell, as shown in Fig.~\ref{fig:p250-identification-validation}, with performance in Table~\ref{tab:jet-identification-errors}. \looseness=-1

\begin{table}
    \caption{Identification errors.}
    \centering
    \begin{tabular}{cc}
        \midrule
        Error                   & Value        \\
        \midrule
        \rowcolor{gray!15}
        Mean Absolute Error     & 4.995~\si{N} \\
        Root Mean Squared Error & 8.061~\si{N} \\
        \bottomrule
    \end{tabular}
    \label{tab:jet-identification-errors}
\end{table}

The model is then embedded in the simulation environment, and the policy is trained as in scenarios described in Sec.~\ref{sec:terrain-aware-locomotion}. The policy outputs the throttle input $u$, which is then passed to the LSTM model~\eqref{eq:jet-dynamics-lstm} along with the previous thrust $T[k-1]$ to obtain the thrust $T$.
The response dynamics of the jet propulsion system exhibit a comparatively slower response rate when compared to the ideal thrust case: attaining the take-off thrust requires a longer time.
For this reason, the policy is trained by reducing the thrust penalization to $1e-8$. Furthermore, the minimum throttle input is set to $15\%$, which leads to a minimum thrust of $\sim40~\si{N}$. The policy can learn how to transition between walking and flying, although the transition is slower than the ideal thrust case, e.g., when traversing the stepping stones, the robot employs the aerial locomotion mode.



\begin{table}
    \caption{Comparison between the policy and the TO method.}
    \centering
    \tabular{ccc}
    \toprule
    \multirow{2}*{Feature} & \multicolumn{2}{c}{Method}             \\
    \cmidrule(lr){2-3}
    & Our & TO   \\
    \midrule
    \rowcolor{gray!15}
    Computation time & $~\simeq 2$ hours & $\simeq 30$ min \\
    Length & Episode length & 7 seconds \\
    \rowcolor{gray!15}
    Cost function & 4 terms & 12 terms \\
    Terrain-aware & \cmark & \xmark \\
    \rowcolor{gray!15}
    Online & \cmark & \xmark \\
    Automatic Transition & \cmark & \cmark \\
    \bottomrule
    \endtabular
    \label{tab:comparison}
\end{table}

\section{Conclusions}
\label{sec:conclusions}

Our paper presents a method that enables aerial humanoid robots to seamlessly transition between walking and flying modes.
The proposed strategy leverages the concept of Adversarial Motion Priors to learn a natural gait pattern from human-like gaits and an efficient aerial locomotion pattern from motions obtained using trajectory optimization.

The robot can traverse complex terrains, switching automatically between both locomotion forms, without an explicit constraint on the form of navigation. Although our method has been tested in simulation environments and necessitates further investigations of its reliability in real-world domains, this result marks progress towards the application of aerial humanoid robots in various fields, such as search and rescue and monitoring missions, where diverse and demanding scenarios are encountered.

Looking forward, the integration of more motion priors might broaden the potential of aerial humanoid robotics in future challenges.

\bibliography{references}
\bibliographystyle{IEEEtran}

\end{document}